\documentclass[10pt,twocolumn,letterpaper]{article}

\usepackage{iccv}
\usepackage{times}
\usepackage{epsfig}
\usepackage{graphicx}
\usepackage{amsmath}
\usepackage{amssymb}
\usepackage{multirow}
\usepackage{authblk}

\usepackage[breaklinks=true,bookmarks=false]{hyperref}
\hypersetup{colorlinks=true}

\iccvfinalcopy

\begin{document}

\title{To Learn or Not to Learn: \\ Analyzing the Role of Learning for Navigation in Virtual Environments}

\author[1, 2]{Noriyuki Kojima\thanks{kojimano@umich.edu}}
\author[2]{Jia Deng\thanks{jiadeng@cs.princeton.edu}}

\affil[1]{University of Michigan, Ann Arbor}
\affil[2]{Princeton University}

\maketitle

\begin{abstract}
In this paper we compare learning-based methods and classical methods for navigation in virtual environments. We construct classical navigation agents and demonstrate that they outperform state-of-the-art learning-based agents on two standard benchmarks: MINOS and Stanford Large-Scale 3D Indoor Spaces. We perform detailed analysis to study the strengths and weaknesses of learned agents and classical agents, as well as how characteristics of the virtual environment impact navigation performance. Our results show that learned agents have inferior collision avoidance and memory management, but are superior in handling ambiguity and noise. These results can inform future design of navigation agents. 
\end{abstract}

\section{Introduction}
Navigation is a fundamental task for general intelligence. Humans can easily move from A to B when B's location is geometrically stated (e.g., B is 20m toward the north of A). But the task remains challenging for robots as it requires solving multiple sub-tasks and integrating their solutions including perceiving 3D structure, estimating ego-motion, planning routes, and handling uncertainty of various kinds. 
    
Traditionally, solutions to these sub-tasks are manually hand-coded by humans, and the hand-coded modules are commonly defined as mapper, localization, planner and controller~\cite{thrun2005probabilistic}. These modules are then assembled into navigation pipelines by humans. Such classical approaches have been studied for decades in both Robotics and Computer Vision, and have been used widely for real-world robot deployment~\cite{davison1998mobile, desouza2002vision, thrun2005probabilistic}. Nevertheless, classical methods are still far away from the robustness and accuracy required for complex real-world scenarios. 
    
Inspired by its success in many domains of AI, deep learning has emerged as a promising alternative to classical methods for navigation~\cite{dosovitskiy2016learning, gupta2017cognitive, gupta2017unifying, jaderberg2016reinforcement, parisotto2017neural}. Deep learning is attractive in that with sufficient data, effective solutions can emerge with minimal hand-engineering. Such data-driven solutions can exploit regularities and discover strategies that are superior to hand-coded rules. As success of deep learning in object recognition has shown, state-of-art accuracy can be achieved by training a single network end-to-end without the decomposition into subtasks conventionally thought necessary. Indeed, recent work~\cite{gupta2017cognitive, jaderberg2016reinforcement, parisotto2017neural} has started to demonstrate good performance of such learned agents for navigation in simulated environments. 

\begin{figure}
\begin{center}
\includegraphics[scale=0.40,trim={0cm 0cm 0cm 0cm}]{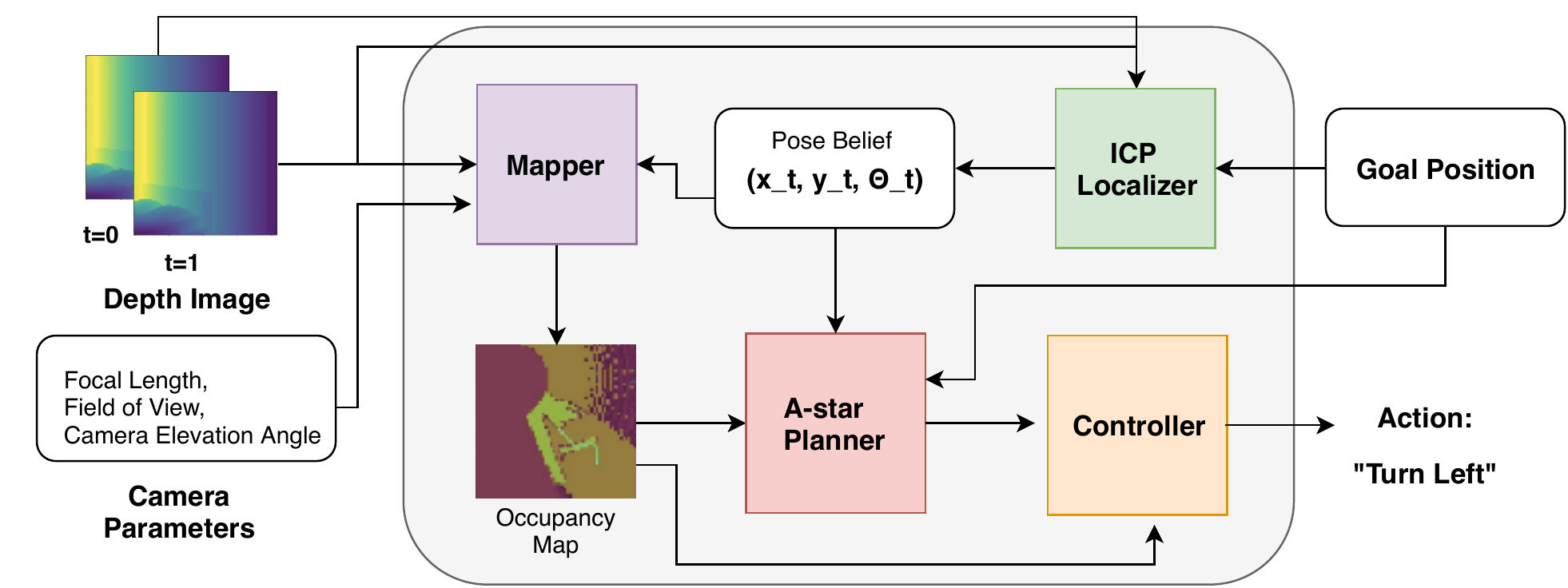}
\end{center}
   \caption{\textbf{Classical Navigation Agent:} We construct a classical navigation agent that consists of a mapper, a localizer, a planner and a controller. See Sec~\ref{ssec:classical_method} for more details.}
\label{fig:classical_navigation_pipelines}
\end{figure}

However, despite these promising results, existing work on learned agents has mostly focused on improving learned agents over learned baselines and has done little or no comparison with classical agents. This leaves many questions unanswered, especially regarding how learned agents stack against classical agents in terms of not only overall performance but also their comparative strengths and weaknesses. In addition, since learned agents have often been tested in virtual environments, it is unclear how challenging these environments are and how their characteristics impact performance. Studying these questions can provide insights not only for building better agents but also for building better testing environments.

In this paper, we take some initial steps toward answering these questions. 
We construct classical navigation agents for two standard virtual environments, MINOS~\cite{savva2017minos} and Stanford Large-Scale 3D Indoor Spaces (S3DIS)~\cite{armeni20163d}, and compare them directly against recent learned agents including UNREAL~\cite{jaderberg2016reinforcement} and Cognitive Mapper and Planner (CMP)~\cite{gupta2017cognitive} on the task of navigating to a goal position with depth input. We show that the classical agents significantly outperform the learned agents.

Next, we investigate the strengths and weaknesses of each method. Previous work has reported only high-level evaluation: success rates / SPL~\cite{anderson2018evaluation} or execution time. High-level metrics are limited as they do not reveal why and how agents fail. Hence, we propose metrics to diagnose various aspects of navigation capability---collision avoidance, memory management, and exploitation of available information. These metrics provide insight into the strengths and weaknesses of each agent. Our analysis shows that learning-based methods, compared to classical ones, have inferior perception and memory management while learning-based methods can better exploit the regularities of the environment and are more robust to noise in perception.  

Finally, we study how characteristics of an environment impact navigation performance. We propose two metrics, ambiguity and complexity, to quantify the intrinsic difficulty of an environment. Ambiguity reflects the difficulty due to partial observation of the environment. Complexity reflects the difficulty due to  the level of physical maneuvering required to reach the goal. We show that learned agents are superior in handling ambiguity but inferior in handling complexity. 

To summarize, our contributions are twofold. First, we construct a classical agent and compare it to learning-based ones on two navigation benchmarks. Second, we offer detailed analysis on the strengths and weaknesses of learned methods and classical methods, as well as how characteristics of environments impact navigation performance. We release source code (\href{https://github.com/princeton-vl/navigation_analysis}{https://github.com/princeton-vl/navigation\_analysis}) to facilitate future research. 

\section{Related Work}
\smallskip \noindent \textbf{Error analysis in learning for navigation:}
Error analysis has played an important role in computer vision research such as object detection~\cite{hoiem2012diagnosing} and VQA~\cite{kafle2017analysis}. Although many learning-based methods have recently been proposed for navigation~\cite{dosovitskiy2016learning, gupta2017cognitive, gupta2017unifying, jaderberg2016reinforcement, parisotto2017neural}, there has been little work focused on error analysis of state-of-the-art methods. The closest to ours is the concurrent works by Mishkin \etal~\cite{mishkin2019benchmarking} and Savva \etal~\cite{habitat19arxiv}, who bench-marked learned agents against classical ones in indoor simulators. Our work shares similarity in comparing learned and classical agents, but is different in that we propose new metrics to diagnose various aspects of navigation capability including collision, avoidance, memory management, and exploitation of available information. 

\smallskip \noindent \textbf{End-to-end learning for navigation}
End-to-end learning for navigation has mostly taken the form of deep reinforcement learning (deep RL)~\cite{dosovitskiy2016learning, mnih2016asynchronous, mnih2016asynchronous, parisotto2017neural, pathak2017curiosity}. From asynchronous training~\cite{mnih2016asynchronous} to curiosity-driven exploration~\cite{pathak2017curiosity}, the network architectures proposed have been quite generic without special design that incorporates domain knowledge from classical navigation research. In our work, we analyze  UNREAL~\cite{jaderberg2016reinforcement}, a state-of-the-art deep RL-based approach for navigation. UNREAL includes a CNN-LSTM based architecture trained through asynchronous auxiliary training. 

Another line of research follows a more module approach by developing learning-based navigation modules which can be integrated to a larger network. For example, localization can be formulated as a 3-DOF or 6-DOF camera pose estimation problem and performed by a deep network~\cite{brachmann2017dsac,kendall2015posenet,melekhov2017relative}. Learning-based approaches have also been studied in the context of SLAM~\cite{bloesch2018codeslam, tateno2017cnn, zhang2017neural}.  Most relevant to our work, Tamar \etal propose Value Iteration Network (VIN)~\cite{tamar2016value} as a differential planner, and Gupta \etal integrate VIN with a differential mapper and propose CMP in~\cite{gupta2017cognitive}, an end-to-end mapper-planner which we analyze as the state-of-the-art method with specially designed components.  

Along with the development of learning methods, there has also been a proliferation of indoor virtual environments~\cite{armeni20163d, brockman2016openai, kempka2016vizdoom, konstantinova2013interactive, savva2017minos, xia2018gibson, yan2018chalet}. These environments provide low-cost platforms for developing navigation methods, and the majority of learning-based navigation methods are studied using these virtual environments. Since virtual environments vary in terms of perceptual input, scenes, and physics engine, evaluating methods across many different environments is impractical. For our study we focus on MINOS~\cite{savva2017minos} and S3DIS~\cite{armeni20163d}, two popular environments for indoor navigation. 

\smallskip \noindent \textbf{Classical Navigation Methods:}
Classical navigation pipelines have been studied for decades. SLAM based approaches~\cite{davison1998mobile, desouza2002vision, thrun2005probabilistic} are commonly used variants of classical methods. SLAM based methods decompose the task into two subtasks---(1) localization and mapping of environments and (2) planning and controlling actions in the constructed map. The first subtask is commonly studied with depth information~\cite{davison1998mobile, minguez2006metric, mur2015orb, thrun2005probabilistic}. In our study, following previous work, we construct classical navigation agents consisting of separate modules for localization, mapping, planning and control. We benchmark our classical agents in MINOS and S3DIS and compare them to learning-based methods.

\section{Evaluation Setup}\label{ssec:tasks}

\noindent \textbf{MINOS:} MINOS is a virtual environment to simulate 3D indoor scenes in Matterport3D~\cite{chang2017matterport3d} and SUNCG~\cite{song2017semantic}. In our experiments, we use synthetic houses from SUNCG. MINOS is integrated with a physics simulator that simulates friction and collision in the environment, i.e., agents slip during actuation, and they get push-backs from objects in collisions. 

We use the same setup from~\cite{savva2017minos}. Houses in MINOS are continuous state spaces, and an agent chooses one of three possible actions at each step: "Move forward by 0.2m", "Rotate left by 7.2 degrees" and "Rotate right by 7.2 degrees". Here 0.2 are 7.2 are fixed constants. At each step, an agent is provided with (1) a depth image and (2) the distance and direction to the goal relative to the current view, i.e.\@ ground truth localization is not provided. 

At the beginning of each episode, an agent is initialized at a random location in a house, and the episode is successful if the agent can reach within 0.2 m of the goal in 500 steps. In MINOS, the houses are grouped into small (2 rooms) and medium (3 to 5 rooms) and are evaluated as two separate tasks. There is a total of 49 small houses and 151 medium houses, with a train-val-test split of 33-7-9 and  87-32-32 respectively. Same as~\cite{savva2017minos}, we use 100 presampled episodes per house for training, 10  for validation, and 10 for test, as provided by MINOS. 

Note that these presampled episodes were selected by MINOS to contain only  nontrivial episodes, with the constraint that to reach the goal, the agent must go through at least one door in a small house and two doors in a medium house. However, the initial version of the open source code of MINOS did not enforce this constraint (and thus included many ``trivial'' episodes), but the current version does.  We believe that the performance numbers of the UNREAL agent reported in~\cite{savva2017minos} were generated with these trivial episodes. This is consistent with the fact that our UNREAL-RGB numbers on MINOS in Tab.~\ref{table:MINOS_general} are significantly lower. 

\noindent \textbf{Stanford Large-Scale 3D Indoor Spaces (S3DIS):} 
S3DIS is a realistic 3D indoor environment consisting of scanned scenes, and it contains the six scenes, each being a floor in a building. S3DIS does not include a physics simulator so there is no push-back from objects and no slipping of agents. 

To set up the tasks in S3DIS, we follow the same procedure from~\cite{gupta2017cognitive}. Each floor in S3DIS is treated as a discrete state space, and discretized as a grid of 0.8m by 0.8m square cells. In S3DIS, an agent chooses from four actions at each step: "Move forward by 0.8m", "Rotate left by 90 degrees", "Rotate right by 90 degrees" and "Stay". At each step of an episode, an agent is provided with (1) a depth image and (2) its absolute pose as well as the goal location, i.e., ground truth localization is provided at each step. 

At the beginning of an episode, an agent is initialized at the random location in a house, and the episode is successful if the agent can reach the goal in T steps, where T is the time budget. There are two types of tasks: (1) the goal is within 32 steps away and (2) the goal is within 64 steps away. Two time budgets (T=39 and T=79) are evaluated for each task. Following~\cite{gupta2017cognitive}, we do a 4-1-1 train-val-test split, and use the exact same sampled episodes.

\begin{figure}[t]
\begin{center}
\includegraphics[scale=0.60, trim={8cm 0cm 8cm 0cm}]{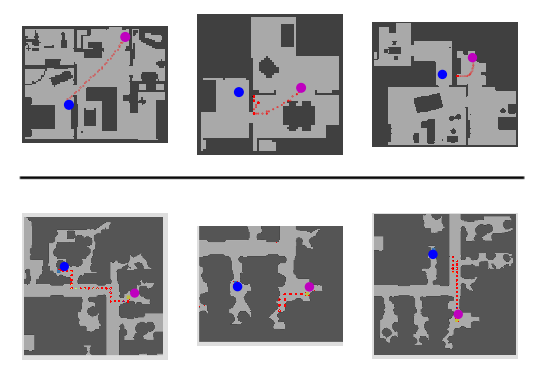}
\end{center}
   \caption{\textbf{Success and failure examples in MINOS and S3DIS:} We visualize example success and failure cases of UNREAL and CMP. Blue dots and magenta dots in figures are the goals and starts respectively. Red dots are trajectories of the agents. \textbf{Top row}: Trajectories of UNREAL in MINOS. The left image is a success episode, the middle image is a failed episode due to collisions, and the right image is a failed episode due to short-term thrashing. \textbf{Bottom row}: Trajectories of CMP in S3DIS. The left image is a success episode, the middle image is a failed episode due to short-term thrashing, and the right image is a failed episode due to ambiguity of the map.}
\label{fig:fail_mode_agent}
\end{figure}

\section{Constructing Classical Agents} 
\label{ssec:classical_method} 
Following the common practice of classical navigation pipeline construction~\cite{thrun2005probabilistic}, we hand-code navigation agents consisting of four components---mapper, localizer, planner, and controller as illustrated in Fig.~\ref{fig:classical_navigation_pipelines}. We construct two agents, one for MINOS and one for S3DIS; this is necessary to handle the two different setups as described in Sec~\ref{ssec:tasks}.

\smallskip \noindent \textbf{Localizer:} 
The localizer estimates an agent's translation and rotation from previous steps. This step is necessary to update the belief of the agent's current pose because due to the physics simulation in MINOS, the agent cannot accurately estimate its location based on its motor command. We adopt the Libpointmatcher~\cite{Pomerleau12comp} implementation of the Iterative Closest Point (ICP) algorithm~\cite{zhang1994iterative}---a commonly used filter-based SLAM approach---to predict translation and rotation. If ICP produces an unreasonable estimate (see appendix), we use the median outcome of an action from training data to approximate translations and rotations of agents. Note these priors are rough approximations since the effects of physics are not considered. In S3DIS, the ground truth translation and rotation is given from the environment.

\smallskip \noindent \textbf{Mapper:} 
The mapper updates a 2D occupancy map using the current depth image and the belief of the agent's current pose. The 2D occupancy map is a grid, where each cell is associated with a score of being an obstacle.  In both MINOS and S3DIS, we have two mappers:  an analytic mapper and a collision mapper.

The analytic mapper takes current depth image and performs the following five steps: (1) construct a 3D point cloud by projecting the depth image into 3D space; (2) transform the 3D point cloud to align with the 2D occupancy map, according to the agent's current pose belief; (3) classify each 3D point as obstacle or free space by thresholding its height; (4) use a custom rule (see appendix) to update the occupancy score of each cell by its number of 3D points classified as obstacle or free space. 

The collision mapper updates the 2D occupancy map when obstacles are detected through collision. The collision mapper is necessary because some obstacles are out of view. We detect collisions by detecting unexpected outcomes of  translation actions. 
For example, if we take a translation action to go from A to B, and end up  in A (i.e., collisions in S3DIS) or in a location away from B (i.e., collisions in MINOS), B is then deemed an obstacle. 

\smallskip \noindent \textbf{Planner:} 
The planner calculates a path to the goal using the 2D occupancy map and generates a set of waypoints on the path. Our Planner converts the 2D occupancy map into a graph by turning each cell into a node connected to its eight neighbors. We use a custom rule (see appendix) to assign edge weights based on the occupancy score of each cell. We run a weighted A-star search on the graph to calculate a path from the belief of the current position to the goal, and set the nodes on the path as waypoints. 

\smallskip \noindent \textbf{Controller:} 
The controller generates actions of the agent. In MINOS, the controller takes in the 2D occupancy map and the waypoints generated by the planner. It generates a sub-goal by finding the farthest waypoint reachable without hitting obstacles in a straight line from the current position. This step helps simplify the path given by the planner---the path is constrained to a 2D grid but MINOS has a continuous state space so the path can be more complex than necessary. In S3DIS, which has a discrete state space, we choose the closest waypoint as the sub-goal. 
In both MINOS and S3DIS, if the agent is facing toward the sub-goal direction (within a threshhold angle), we take the forward translation action; otherwise we turn left or right  to align the direction.

\begin{table}
\begin{center}
\scalebox{0.85}{
 \begin{tabular}{ c c c c c} 
 \hline
 Metric & \multicolumn{2}{c}{Success Rate} & \multicolumn{2}{c}{Avg \# Steps}\\
 \cline{1-3}\cline{4-5}
 House Size & Small  & Medium & Small & Medium    \\ 
 \hline\hline
 \multicolumn{1}{l}{UNREAL} & 50.5 & 20.2 & 319.4 & 387.6 \\ 
 \multicolumn{1}{l}{Classical} & \textbf{84.3}  & \textbf{77.9} &\textbf{178.3}  &\textbf{255.3}\\
  \hline
  \multicolumn{1}{l}{UNREAL-RGB}  & 28.6 & 16.7 & 379.5 & 447.0\\
  \multicolumn{1}{l}{Classical-GT} & 97.0 & 94.1 & 95.3 & 179.5 \\
 \hline
\end{tabular}
}
\end{center}
\caption{\textbf{MINOS results}: For UNREAL and the classical method, we report the success rate and average number of steps per episode. UNREAL-RGB denotes UNREAL with RGB input; Classical-GT denotes the classical navigation method provided with groundtruth localization. }
\label{table:MINOS_general}
\end{table}

\begin{table}
\begin{center}
\scalebox{0.85}{
 \begin{tabular}{ c c c c c} 
 \hline
 Metric & \multicolumn{2}{c}{Success Rate} & \multicolumn{2}{c}{Mean Distance}\\
 \cline{1-5} 
 Time Budget & 39 steps & 69 Steps  & 39 steps & 69 Steps \\ 
 \hline\hline
 \multicolumn{1}{l}{~~CMP~\cite{gupta2017cognitive}} & 78.8 & 91.5 & 4.8 & \textbf{1.4} \\ 
 \multicolumn{1}{l}{~~Classical} & \textbf{82.8 }& \textbf{95.5} & \textbf{3.1} & 2.1 \\
 \hline
\end{tabular}
}
\end{center}
\caption{\textbf{S3DIS 32 steps task results:} For CMP and the classical method, we report the success rate and mean distance to the goal under the standard time budget (39 steps) and extra time budget (69 steps).}

\label{table:S3DIS_32}
\end{table}
\begin{table}
\begin{center}
\scalebox{0.85}{
 \begin{tabular}{ c c c c c} 
 \hline
 Metric & \multicolumn{2}{c}{Success Rate} & \multicolumn{2}{c}{Mean Distance}\\
 \cline{1-5} 
 Time Budget & 79 steps & 159 Steps  & 79 steps & 159 Steps \\ 
 \hline\hline
 \multicolumn{1}{l}{~~CMP~\cite{gupta2017cognitive}} & 66.3 & 78.5 & 11.9 & 9.3 \\ 
 \multicolumn{1}{l}{~~Classical} & \textbf{70.0}& \textbf{93.6} & \textbf{7.6} & \textbf{3.1} \\
 \hline
\end{tabular}
}
\end{center}
\caption{\textbf{S3DIS 64 steps task results:} For CMP and the classical method, we report the success rate and mean distance to the goal under the standard time budget (79 steps) and extra time budget (159 steps).}
\label{table:S3DIS_64}
\end{table}

\section{Learning-based Navigation Methods}
\noindent \textbf{UNREAL:} 
UNREAL~\cite{jaderberg2016reinforcement} is a state-of-the-art deep reinforcement learning agent for navigation. UNREAL uses a fairly generic CNN-LSTM architecture to process input and predict actions. 

UNREAL is trained with a version of the asynchronous advantage actor-critic method~\cite{wang2016sample}, augmented with auxiliary self-supervised learning. 
In our experiments, we use the same architecture and training objectives as in~\cite{jaderberg2016reinforcement}.  We train and test UNREAL in MINOS environment. We train UNREAL asynchronously with 4 workers up to 13.2M steps using the train episodes of MINOS. We follow the same training procedure as in ~\cite{savva2017minos} except we also train a version with only depth images as input in addition to a version with only RGB.

\smallskip \noindent \textbf{Cognitive Mapper and Planner (CMP):}
CMP~\cite{gupta2017cognitive} is an end-to-end learning-based navigation method with specially designed navigation modules, and the state-of-the-art agent in S3DIS environment. CMP has a mapper module and a planner module, both of which are differentiable. 
CMP is trained through imitation learning using demonstrations. In our experiments, we use the pre-trained model in S3DIS provided by Gupta \etal~\cite{gupta2017cognitive} and test it in S3DIS.

\section{Comparison on Overall Performance}
We compare the overall performance of learning-based and classical methods on MINOS and S3DIS. 

For MINOS, Tab~\ref{table:MINOS_general} reports the success rate and average number of steps. Since Deep RL based methods have a large variance in training, we report the best score for UNREAL from five separate training runs. We can see that our classical agents outperform UNREAL by large margins for all tasks and metrics. In particular, classical agents have an almost 4X higher success rate in the medium-house task. For reference, we also report the performance of UNREAL with RGB input and classical navigation pipeline given groundtruth localization. 

For S3DIS, we report the success rate and the mean distance (number of steps) to the goal at the end of each episode (distance is zero if the goal is reached at the end of the episode) in Tab~\ref{table:S3DIS_32} and Tab~\ref{table:S3DIS_64}. 
In addition to the standard time budget (39 for the 32 steps task and 79 for the 64 steps task) as defined in Sec~\ref{ssec:tasks}, we also report the performance of the agents when they are given extra time:  69  for the 32 steps task and 159 for the 64 steps task. We can see that the classical navigation agent outperforms CMP in 7 out of the 8 metrics. By comparing the performances under the standard time budget and the extra time budget for the same agent, we can see the standard step budget is too tight for both CMP and the classical navigation method. We hypothesize this is due to the ambiguity caused by partial observation of the environment, and we discuss this in Sec~\ref{sec:path_ambiguity}. In addition, when the tasks become more challenging with goals being further away (from 32 steps to 64 steps), CMP suffers from a much bigger drop of performance than the classical agent. Finally, for the more difficult task (the 64 steps task), the performance gap between CMP and the classical agent enlarges as more time is allowed. 

\section{Fine-grained Comparisons}
While comparison of overall performance is useful, it does not inform us of why and how each agent fails and how to improve them. We thus probe deeper into how different agents behave. 

To successfully navigate through environments, an agent needs several skills. First, it needs to perceive and recognize free space and obstacles to avoid collision. Second, it needs to maintain and update their memory to avoid thrashing (i.e., repeatedly visiting the same place). In addition, it needs to exploit its knowledge of the environment to plan actions. Finally, it needs to be able to handle uncertainty and noise from perception. Fig.~\ref{fig:fail_mode_agent} illustrates common failure modes of learning-based methods. We hypothesize these failure modes are related to deficiencies in one of the necessary skills. 

We thus propose metrics to quantify each aspect and compare learned agents and classical ones. We conduct our study using the validation episodes of MINOS and S3DIS (32 steps task). 

\smallskip \noindent \textbf{Collision avoidance} 
Perception  of free spaces and obstacles are necessary to  avoid collision. Therefore, we quantify this ability by studying the frequency of collision in each navigation episode. Because rotational actions do not cause collisions, for an agent we calculate the frequency of collision of its "move forward" action in each episode---the number of collisions divided by the number of "move forward" actions---and average it over all episodes. 

In Tab.~\ref{table:Normalized_Collsion_Measure} we can see both CMP and UNREAL collide with obstacles more frequently than their classical counterpart. In particular, UNREAL has significantly weaker collision avoidance: more than 60$\%$ of their translation actions cause collisions. It is worth noting that UNREAL is trained end-to-end with a generic architecture designed with the least amount of prior knowledge about navigation. This suggests that end-to-end training of UNREAL has failed to make up for the lack of hand-coded prior knowledge.

\begin{table}
\begin{center}
\scalebox{0.85}{
 \begin{tabular}{ c c c c } 
 \hline
 \multirow{2}{*}{Methods} & \multicolumn{3}{c}{Collision Frequency} \\
 \cline{2-4}
 &  MINOS(S) & MINOS(M) & S3DIS   \\ 
 \hline\hline
 \multicolumn{1}{l}{UNREAL}  & 66.1 & 65.3  & - \\ 
 \multicolumn{1}{l}{CMP} & - & - & 2.0 \\ 
\multicolumn{1}{l}{Classical} & \textbf{5.7}  & \textbf{5.5} & \textbf{0.2} \\
  \hline
\end{tabular}
}
\end{center}
\caption{Collision frequency on MINOS and the S3DIS 32 steps task. MINOS(S) denotes the small house task, and MINOS(M) denotes the medium house task.}
\vspace{-2ex}
\label{table:Normalized_Collsion_Measure}
\end{table}

\subsection{Thrashing}
An agent has a higher chance of reaching the goal by exploring different states, rather than getting stuck in the same state or going in circles. Here we quantify how often an agent gets stuck, either locally by repeatedly hitting an obstacle (short-term thrashing) or traveling in loops (long-term thrashing). Thrashing can be avoided by proper memory management that remembers past mistakes and avoids repeating them. We measure short-term thrashing frequency $Th_s$ and long-term thrashing frequency $Th_l$. Short-term thrashing evaluates how often the agent applies the same action immediately after the previous action caused a collision, and long-term thrashing evaluates how often the agent revisits a location that has been visited before. 

We use the following notations for the formal definition of $Th_s$ and $Th_l$: $a_t$ is the action performed at step $t$; $c_t$ is a binary value to indicate whether there is a collision at step $t$; $l_t$ is the 2D coordinates at step $t$; $T$ is the number of steps taken in the episode. We also define the function $\text{fn}(l_1, l_2, \text{th}) = [dist(l_1, l_2) > th]$, which returns 1 if the distance of two locations is more than the threshold $\text{th}$, and 0 otherwise. We formally define  $Th_s$ and $Th_l$ below: 

\begin{equation*} 
\label{Point Clouds}
\begin{split}
Th_s = 100*\frac{\sum^{T-1}_{t=0} c_t* [a_t == a_{t+1}]}{\sum^{T-1}_{t=0} c_t}
\end{split}
\end{equation*}

\begin{equation*}
\small
\begin{split}
Th_l = 100 * (1 - \frac{\sum^{T-1}_{t=0} \text{fn}(l_t, l_{t+1}, \delta) * \prod^{t-1}_{t_2=0} \text{fn}(l_t, l_{t_2}, \epsilon) }{\sum^{T-1}_{t=0} \text{fn}(l_t, l_{t+1}, \delta)}).
\end{split}
\label{long-term-thrashing}
\end{equation*}
We set $\delta$ to be a small positive number, and $\epsilon$ to be the step size of the agents in the environment (0.2 m in MINOS and 0.8m in S3DIS.) 

We calculate $Th_s$ and $Th_l$ for each episode and report the mean in Tab~\ref{table: thrashing measures}. The results suggest that CMP wastes over 20\% on short term thrashing and UNREAL around 50\% to 60 \%, whereas there is almost negligible short term thrashing for the classical agents. In addition, we see that both UNREAL and CMP suffer more from long term thrashing than the classical agents, that is, they are more likely to travel in circles. These results show that  learning-based methods are less capable of recognizing and avoiding previously visited states, short-term or long-term, suggesting more room for improvement in terms of memory management. 

\begin{table}
\begin{center}
\scalebox{0.85}{
 \begin{tabular}{ c c c c } 
 \hline
 \multirow{2}{*}{Methods} & \multicolumn{3}{c}{Thrashing Frequency} \\
 \cline{2-4}
 & MINOS(S) & MINOS(M) & S3DIS  \\ 
 \hline\hline
 \multicolumn{1}{l}{\textbf{Short-term }}& & & \\ 
 \multicolumn{1}{l}{UNREAL}  & 65.6 & 51.1  & - \\ 
 \multicolumn{1}{l}{CMP} & - & - & 21.0 \\ 
 \multicolumn{1}{l}{Classical} & \textbf{2.4}  & \textbf{0.7} & \textbf{0} \\
 \hline
 \multicolumn{1}{l}{\textbf{Long-term}}& & & \\ 
 \multicolumn{1}{l}{UNREAL}  & 59.9 & 49.9  & - \\ 
 \multicolumn{1}{l}{CMP} & - & - & 34.8 \\ 
 \multicolumn{1}{l}{Classical} & \textbf{48.9}  & \textbf{48.3} & \textbf{27.0} \\
 \hline
\end{tabular}
}
\end{center}
\caption{\textbf{Short-term $Th_s$ and long-term $Th_l$ thrashing frequency:} We report mean $Th_s$ and $Th_l$ for MINOS and the S3DIS 32 steps tasks. MINOS(S) denotes the small house, and MINOS(M) denotes the medium house.}
\vspace{-2ex}
\label{table: thrashing measures}
\end{table}

\subsection{Exploitation}
An important aspect of navigation skill is to quickly reach the goal without unnecessary exploration of the environment, that is, the ability to exploit the currently available information to the maximum extent. 

We propose a metric $M_e$ to measure the exploitation ability of an agent. We evaluate the number of average unique 3D bins observed and required to get one step closer to the goal. In each environment, we discretize the global 3D space into 3D bins, and $N_b$ is the number of bins containing 1 or more 3D points observed by the agent during one episode. $D_{i}$ is the distance of the optimal path from the current location to the goal based on the groundtruth 2D map at step i.
\begin{equation*} 
\label{Point Clouds}
\begin{split}
M_e = \frac{N_{b}}{D_{0}-min(D_1, ...,D_T) + 1}\\
\end{split}
\end{equation*}
The denominator of $M_e$ calculates the maximum reduction of distance to the goal achieved by the agent during an episode; we add 1 to avoid division by 0. A lower $M_e$ suggests that the agent has a higher capability of getting closer to the goal with less information. We calculate $M_e$ for individual episodes and report the median (because $M_e$ is not bounded and can be distorted by outliers) in Tab~\ref{Table: Exploitation}. We see that the classical method outperforms UNREAL on this metric. However, CMP outperforms its corresponding classical method, showing that CMP is capable of exploiting the regularities of environment learned from training data. 

\begin{table}
\begin{center}
\scalebox{0.85}{
 \begin{tabular}{ c c c c } 
 \hline
 \multirow{2}{*}{Methods} & \multicolumn{3}{c}{Exploitation Measure} \\
 \cline{2-4}
 &  MINOS(S) & MINOS(M) & S3DIS  \\ 
 \hline\hline
 \multicolumn{1}{l}{UNREAL}  & 608.6 & 882.2  & - \\ 
 \multicolumn{1}{l}{CMP} & - & - & \textbf{16735.3} \\ 
 \multicolumn{1}{l}{Classical} & \textbf{226.8}  & \textbf{300.9} &  18751.8 \\
 \hline
\end{tabular}
}
\end{center}
\caption{\textbf{Exploitation measure $M_e$:} We report mean $M_e$ for MINOS and the S3DIS 32 steps task. MINOS(S) denotes the small house task, and MINOS(M) denotes the medium house task. Note that due to  different discretization of 3D space in MINOS and S3DIS, the scores are not comparable across environments.}
\vspace{-2ex}
\label{Table: Exploitation}
\end{table}

\begin{figure*}[t]
\begin{center}
\includegraphics[scale=1.3,trim={0cm 0cm 0cm 0cm}]{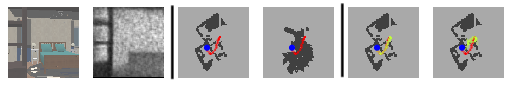}
\end{center}
   \caption{\textbf{Effects of noisy depth on the classical agent in MINOS:} We visualize the effects of depth noise coming from the FCRN depth estimator. For the middle columns and right columns, the dark gray regions are predicted obstacles, and light gray regions are predicted free space. The blue dot is the goal, and the red dots show a trajectory of an agent. \textbf{Left:} an input RGB image and the predicted depth by FCRN. \textbf{Middle:} 2D Maps predicted by the mapper when the optimal path and groundtruth localization are provided. The left map is constructed using groundtruth depth, and the right map is constructed by depth predicted by FCRN. \textbf{Right:} Trajectories predicted by the localizer when the optimal actions are provided. Yellow dots show a trajectory predicted by the localizer. The left trajectory is calculated with groundtruth depth, and the right trajectory is calculated using depth estimated by FCRN.}
\vspace{-2ex}
\label{fig:RGB2Depth}
\end{figure*}

\subsection{Noisy perception}
Finally, we investigate the effects of noisy depth input. This is important to study because unlike a virtual environment, real-world sensors invariably have errors with various causes. In our study, we consider two cases: (1) when the depth images contain Gaussian noise and (2) when the depth images are estimated from RGB.

Our experiments show the learning-based agents have significantly better robustness toward the noisy depth in comparison to the classical navigation method; this observation holds across all of our experiments. The classical navigation agents are not robust to perceptual noise because their design propagates noise from the mapper and localizer to the planner and controller as illustrated in Fig~\ref{fig:RGB2Depth}

\smallskip \noindent \textbf{Gaussian noise:} In this study, we add Gaussian noise to the groundtruth depth images and use these noisy depth images as the input to an agent. We generate pixel-wise Gaussian noise centered at 0 with a standard deviation of $\sigma$. We experiment with different $\sigma$ values to investigate the effects of different noise levels. For example, we define a 50 $\%$ noise level to mean that 99.7 $\%$ of the added noise is less than or equal to half of the step size of the agent. In other words, for a 50 $\%$ noise level, we set $3\sigma$ to be 0.1 m in MINOS and 0.4m in S3DIS.

\begin{figure}
\begin{center}
\scalebox{1}[0.8]{
\includegraphics[scale=0.2,trim={0cm 0cm 0cm 1cm}]{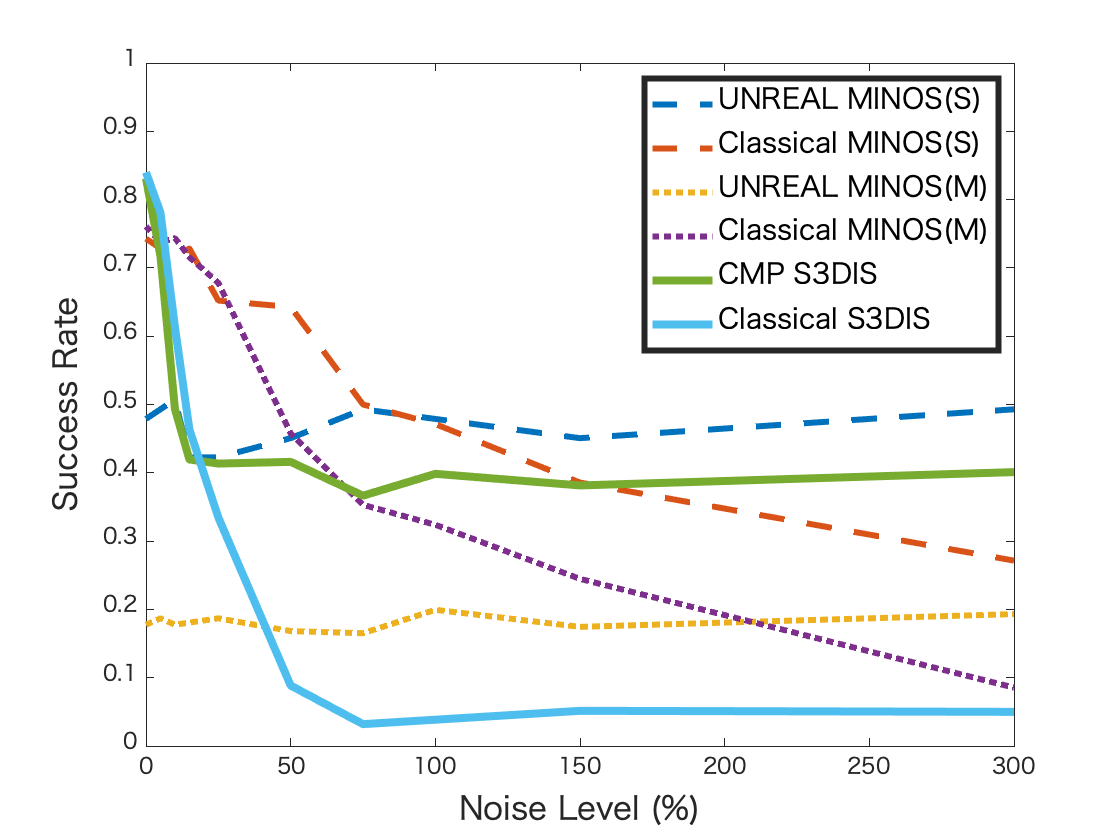}
}
\end{center}
   \caption{\textbf{All methods under Gaussian noise:} We report the success rate of all agents on MINOS and S3DIS under different Gaussian noise levels. We use dot lines, dash lines and solid lines to show results for the validation episodes of MINOS small and medium house, and the S3DIS 32 steps task respectively.}
\label{fig:all_gauss_noise}
\end{figure}

Fig.~\ref{fig:all_gauss_noise} shows our results for all methods in MINOS and the S3DIS 32 steps task. Fig.~\ref{fig:all_gauss_noise}  reports the success rate of all methods with different noise levels. In MINOS, UNREAL  suffers very little from Gaussian noise, while the classical agent is greatly impacted. UNREAL eventually outperforms the classical agent as the noise level increases. In S3DIS, the performance of both CMP and the classical agent immediately start to decrease with only a small amount Gaussian noise added to the clean depth images. However, we can see the success rate of CMP stops dropping as the noise level increases further, while the classical method continues to drop. CMP outperforms the classical method by a margin of more than the 35 $\%$ at certain noise level, demonstrating significantly better robustness to noise of CMP than the classical method.

\noindent \textbf{Depth estimated from RGB:} In this experiment, only RGB images are provided from the environments, and we estimate depth using a depth estimator. We feed the estimated death images to the agent to obtain the navigation performance. We test this set up on UNREAL and the classical  agent on the validation episodes of MINOS. We use FCRN~\cite{laina2016deeper} as our depth estimator. We train FCRN from scratch using the 70k RGB-D images randomly sampled from the scenes in the training episodes. We validate the performance of FCRN on 1k images (sampled from the validation scenes) in MINOS, and we obtained an RMSE of 0.6231.

We report the success rate and the average number of steps per episode for MINOS in Tab~\ref{table:minos_fcrn}. While noise coming from the depth estimator affects the performance of both UNREAL and the classical  agent, we can see the classical method is more severely impacted as shown in Tab~\ref{table:minos_fcrn}, and it gets outperformed by UNREAL.  

\begin{table}
\begin{center}
\centering
\scalebox{0.85}{
 \begin{tabular}{c c c c c} 
  \hline
   \multirow{2}{*}{Methods} & \multicolumn{2}{c}{Success Rate} & \multicolumn{2}{c}{Average Steps}\\
    \cline{2-3} \cline{4-5} & \fontsize{8}{8}{Small} & \fontsize{8}{8}{Medium} & \fontsize{8}{8}{Small} & \fontsize{8}{8}{Medium} \\ 
    \hline\hline
      \multicolumn{1}{l}{UNREAL (FCRN)} & \textbf{40.8} & \textbf{13.7} & \textbf{352.0} & \textbf{454.4} \\ 
 \multicolumn{1}{l}{Classical (FCRN)} & 11.4 & 0.9 & 473.7 &  498.3 \\ 
 \hline
   \multicolumn{1}{l}{UNREAL (Clean-Depth)} & 47.8 & 17.7 & 319.4 & 387.6 \\ 
  \multicolumn{1}{l}{Classical (Clean-Depth)}  & \textbf{74.2}  & \textbf{76.0} & \textbf{205.9}  &\textbf{248.8}  \\

  \hline
 \end{tabular}
}
\end{center}
\caption{\textbf{MINOS with depth estimator from RGB:} We report the success rate and average number of steps per episode for UNREAL and the classical agent on the validation episodes of MINOS. The upper rows show the results when the depth is estimated by FCRN from RGB, and the lower rows show the results with clean depth.}
\label{table:minos_fcrn}
\end{table}

\section{Analysis of Environments}
Current navigation research has mostly focused on studying how to build  agents, and has not devoted much effort into studying the intrinsic characteristics of the environment (e.g.\@ geometric features of environments) and how they impact performance. Analyzing the environment can not only shed light on how to design better environments, but can also help compare agents across environments. For example, if we have two agents---Agent A with an 80 \% success rate by reaching goals through the straight hall-way free of obstacles versus Agent B with a 75 \% success rate by reaching goals through the several complex office rooms---comparing A and B needs to account for the difficulty of each environment. In this section, we analyze an environment through ambiguity and complexity of the optimal path, and we investigate behaviors of learned methods and classical methods along the way.

\begin{figure}[t]
\begin{center}
\includegraphics[scale=0.6,trim={1.5cm 1cm 0cm 0.5cm}]{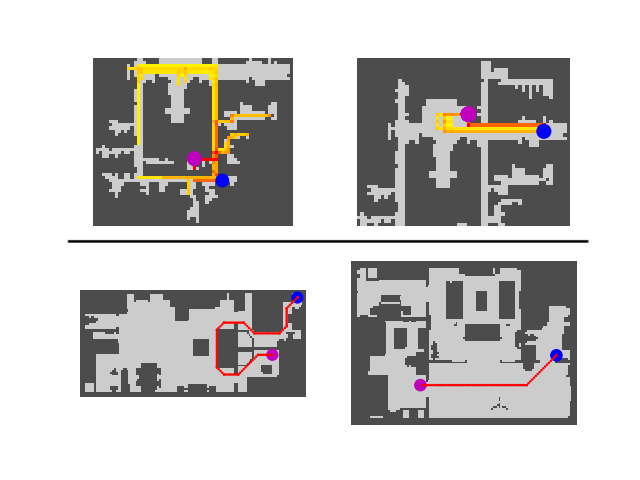}
\end{center}
\vspace{-1ex}
\caption{\textbf{Examples of ambiguity / complexity:} The magenta dot is the start, and the blue dot is the goal. \textbf{Top row:} The left map is an ambiguous task (7.5 ambiguity score) and the right map is an unambiguous task (1.0 ambiguity score). The colors in the figure shows a heatmap of 2D-MC's trajectories. \textbf{Bottom row:} The left map is a  episode with high complexity (10 turns), and the right map is an episode with low complexity (1 turn).}
\vspace{-2ex}
\label{fig:a_c_examples}
\end{figure}

\subsection{Ambiguity} ~\label{sec:path_ambiguity}
An ambiguous environment is one that looks like a maze. That is, path A and B can both plausibly lead to C, but there is no way to infer which one actually leads to C based on available information. By studying ambiguity in an environment, we want to quantify the difficulty resulting from incomplete knowledge of the environment. Furthermore, we study whether the learning-based methods have learned regularities to make better decisions when they encounter ambiguity in the environment, in comparison with a classical agent that does not use any priors about the environment. 

To this end, we construct an idealized 2D Monte Carlo (2D-MC) agent to quantify ambiguity of individual navigation tasks. 2D-MC navigates in a 2D map free of physics. We assume the perfect perception for 2D-MC, and 2D-MC observes all free spaces and obstacles in all directions in the ground truth 2D map. However, 2D-MC cannot see through walls. We define a frontier as the boundary between observed free space and unobserved space. If 2D-MC observes the goal directly, it plans actions toward the goal. Otherwise it randomly sample a point on the frontier and plan actions toward the goal---this is the optimal action without assuming any prior about the environment because there is no reason to believe any frontier point is better than the others in leading to the goal. This process continues until 2D-MC finds the goal. We keep track of the number of actions taken by 2D-MC. We run 2D-MC on each task 20 times, and denote the median as $N_{mc}$. More implementation details of 2D-MC can be found in our appendix.

We calculate the ambiguity score of each episode by simply dividing $N_{mc}$ with the number of actions necessary to trace the optimal path. Intuitively, the ambiguity score measures how much extra effort is caused due to incomplete knowledge. 

To investigate how an agent performs under ambiguity, we calculate the ambiguity score for all the validation episodes from the MINOS medium house and the S3DIS 32 steps task, and we compute correlation coefficients against the agent's SPL~\cite{anderson2018evaluation} calculated for each episode. SPL is the success rate discounted by the number of actions taken in comparison to the optimal path. For S3DIS,   the correlation coefficient is -0.0095 for CMP and -0.1880 for the classical agent. For MINOS,  the correlation coefficient is 0.1349 for UNREAL and -0.4476 for the classical agent. Stronger negative correlation means the agent tends to take extra steps for more ambiguous tasks, suggesting that learning-based methods can better exploit regularities to handle ambiguity than the classical method.

\subsection{Complexity} 
By complexity, we capture how "twisted" the groundtruth path is. By analyzing the path complexity, we aim to measure difficulty caused by the geometric configuration of the environment. The more complex a path is, the more sophisticated the agent needs to be in obstable avoidance and planning. In our study, we use the number of turns in an optimal path to represent the path complexity. 

We calculate the number of turns in the optimal path and an agent's SPL~\cite{anderson2018evaluation} for each validation episode from the MINOS medium house task and the S3DIS 32 steps task. We compute  correlation coefficients between path complexity and SPL to quantify the extent to which agents are susceptible to path complexity. For S3DIS, the  correlation coefficient is -0.1421 for CMP and -0.1229 for the classical method. For MINOS, the correlation coefficient is -0.2361 for UNREAL and -0.2012 for the classical method. Stronger negative correlation coefficients imply that the agent tends to struggle more on twisted paths, suggesting that the classical methods are superior at handling path complexity over learning-based methods.

\section{Conclusions}
We have compared learning-based methods and classical methods for navigation in virtual environments. We have constructed classical navigation agents and demonstrate that they outperform state-of-the-art learning-based agents on two standard benchmarks: MINOS and Stanford Large-Scale 3D Indoor Spaces. We have performed detailed analysis to study the strengths and weaknesses of learned agents and classical agents, as well as how characteristics of the virtual environment impact navigation performance. Our results show that learned agents have inferior collision avoidance and memory management, but are superior in handling ambiguity and noise. These results can inform future design of navigation agents. 
\maketitle
\begin{figure*}
\begin{center}
\includegraphics[scale=0.8,trim={0cm 0cm 0cm 2cm}]{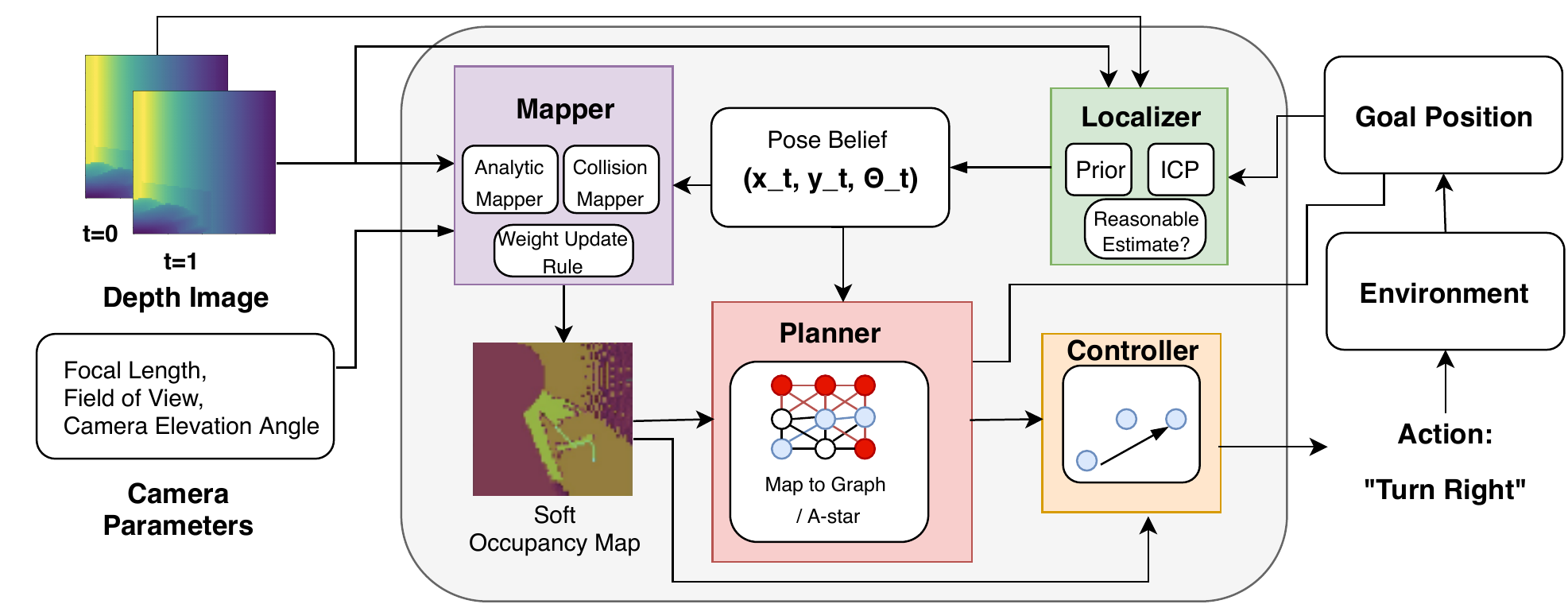}
\end{center}
   \caption{\textbf{Classical Navigation Pipeline in MINOS:} The figure above illustrates the classical navigation pipeline constructed for MINOS environment. We describe details in Section 4 and A1.}
\label{fig:minos_classical_navigation_pipeline}
\end{figure*}

\begin{figure*}
\begin{center}
\includegraphics[scale=0.8,trim={0cm 0cm 0cm 0cm}]{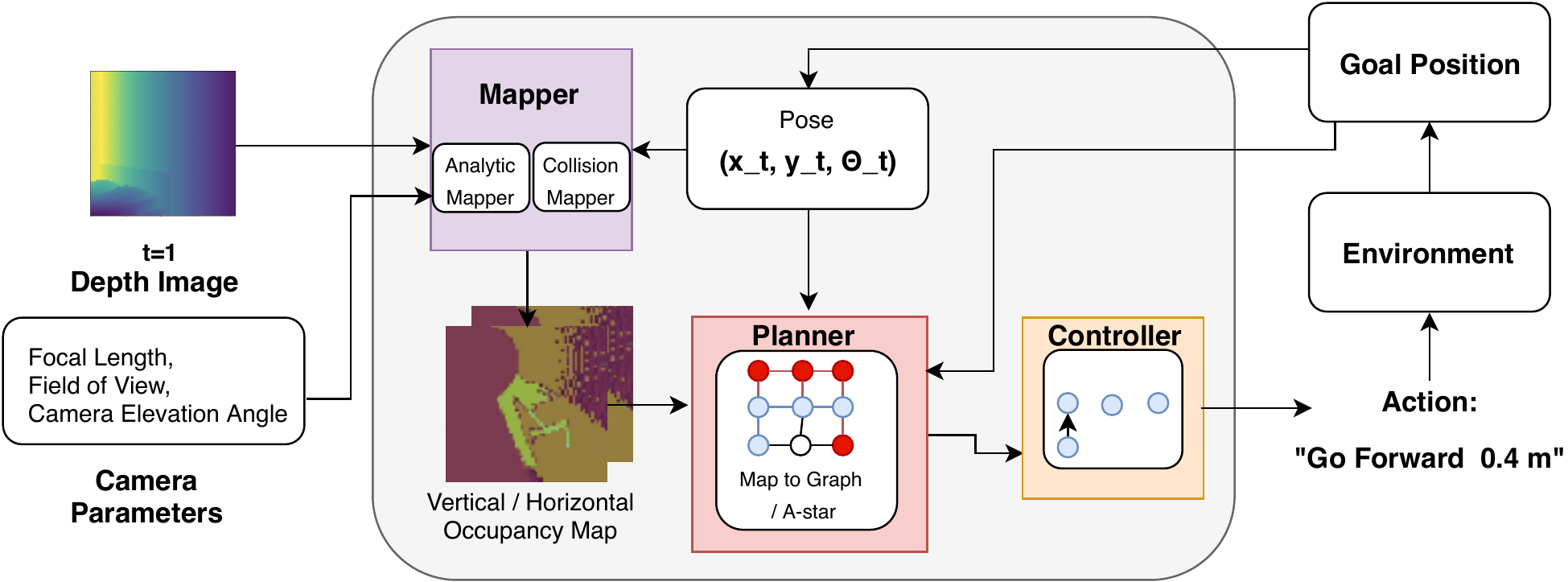}
\end{center}
   \caption{\textbf{Classical Navigation Pipeline in S3DIS:} The figure above illustrates the classical navigation pipeline constructed for S3DIS environment. We describe details in Section 4 and A2.}
\label{fig:s3dis_classical_navigation_pipeline}
\end{figure*}

\begin{figure*}
\begin{center}
\includegraphics[scale=0.8,trim={0cm 0cm 0cm 0cm}]{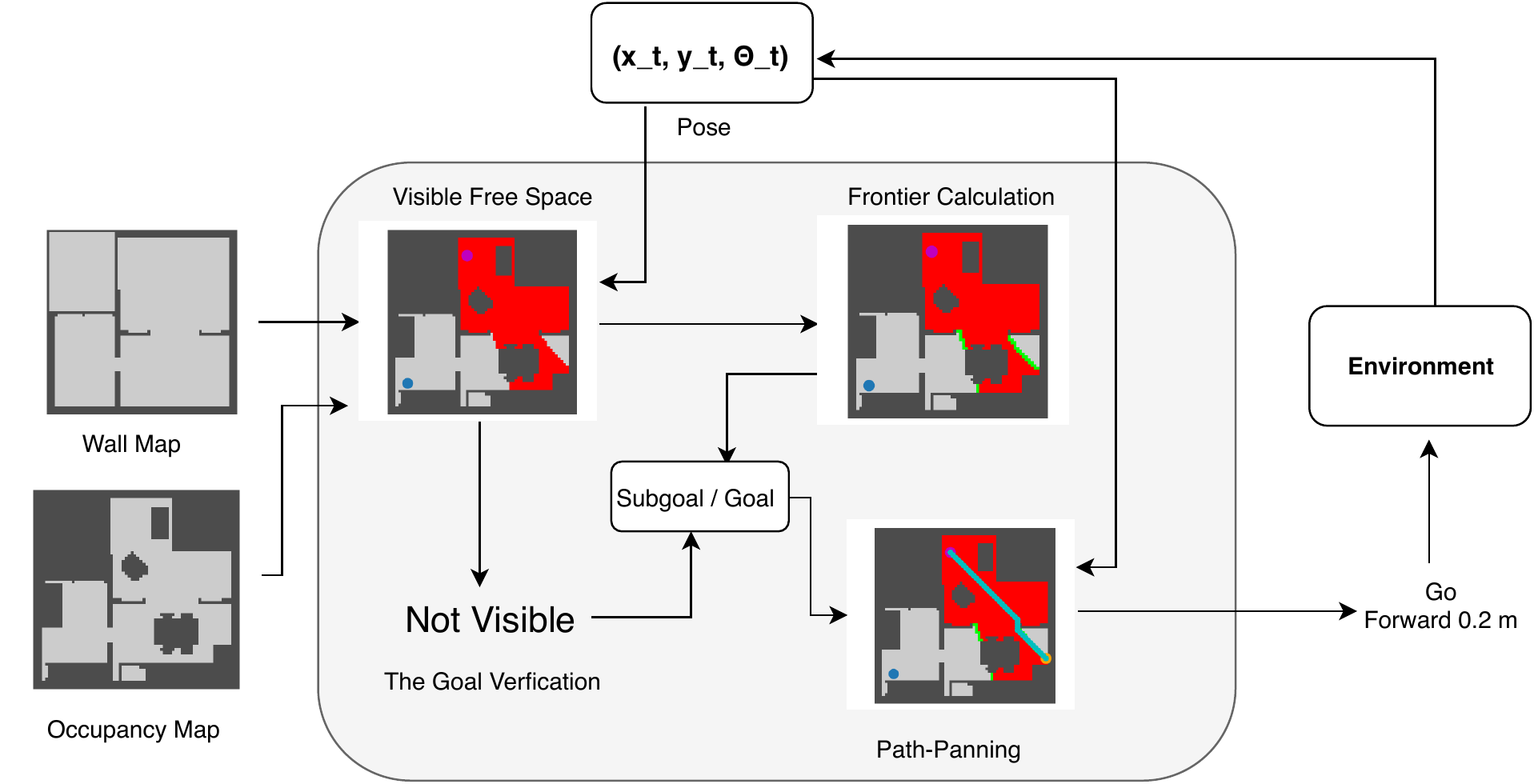}
\end{center}
   \caption{\textbf{Idealized 2D Monte Carlo Agent Pipeline:} The figure above illustrates the 2D-MC agent proposed for experiments in Section 8.1 of the paper. A magenta dot in a map shows the start location, a blue dot indicates the goal location, and an orange dot is a sampled subgoal. Red regions on the maps show the observed free space, and green regions show frontiers. Finally, an emerald line illustrates a planned path to the subgoal. We describe details in Section A3 of the appendix.}
\label{fig:2d-mc_navigation_pipeline}
\end{figure*}

\section*{Appendix}
\subsection* {A1. Classical Navigation Pipeline in MINOS}~\label{sec:path_ambiguity}

\noindent \textbf{Localizer---ICP Configuration:} As mentioned in the previous section, we use the Libpointmatcher~\cite{Pomerleau12comp} implementation of the Iterative Closest Point (ICP) algorithm~\cite{zhang1994iterative} to predict translations and rotations of an agent at each step. Iterative Closest Point is a filter-based SLAM, and it chains together four main components---data filters, a matcher, outlier filters and a minimizer. Given both a current point cloud (i.e., a point cloud at step t) and a previous point cloud (i.e., a point cloud at step t-1), the data filters remove noise and redundancy, and generate descriptive features. Afterward, ICP iteratively (1: matcher) matches points between the two clouds, (2: outlier filters) removes outliers, and (3: minimizer) calculates the transformation of the current cloud on the previous cloud. 

We customize the configuration of ICP---we add additional descriptive features to the point clouds by estimating the surface normal at each 3D point and calculating the orientation of each 3D point from the current viewpoint; We use the KD-Tree algorithm for our matcher and the Point-to-Plane~\cite{chen1992object} optimization for our minimizer. The exact configurations are included in our code release. 

Finally, to improve the robustness of the translations/rotation prediction, we run the ICP algorithm three different times, varying descriptive features and inputs (2D vs. 3D point clouds). Among three possible predictions from ICP, we output the translations and rotation which maximize the consistency (minimize the distance) of two predictions of the agent's current coordinates calculated from different priors and estimation. We discuss this consistency with details in the next section (under ICP Unreasonable Estimation / Location Consistency.)  

\smallskip \noindent \textbf{Localizer---ICP Unreasonable Estimation / Location Consistency:} At the beginning of each episode, we initialize the agent's current coordinates, the goal coordinates and the agent's camera orientation on the 2D occupancy map. Note that the goal coordinates are constant through an episode. We define the consistency to be the difference (distance) of the two estimations made for the agent's current coordinates.

(a) The first coordinates estimation is produced by predicting the agent's translations from step t-1 to t and adding the translations to the current coordinates estimated at step t-1. The prediction of the translations made at each step is produced by ICP in our pipeline. 

(b) The second coordinates estimation is produced from estimating the difference of the coordinates between the agent's current position and the goal, and subtracting the difference from the goal coordinates. We calculate this difference for two axes X and Y by applying trigonometry to the Euclidean distance and orientation to the goal, both of which are provided from the MINOS environment. However, this orientation to the goal only tells the orientation of the goal respect to the current view direction. To align the goal orientation to the frame of the 2D occupancy map, we simply add the current camera orientation of the agent to the orientation to the goal. Note that the agent's current camera orientation is updated at each step from the rotation predicted by ICP.  

Minimizing the distance error between two coordinates estimations (a) and (b) means maximizing the consistency. In our experiments, we set the error threshold to 0.2 m. 

\smallskip \noindent \textbf{Localizer---Fine-tune Translations:} At the last stage of localizer, we fine-tune the translations predicted by ICP. We fix the translations prediction to minimize the error in the consistency discussed above, assuming the rotation predicted by ICP is correct. We make this assumption because the ambiguity in matching two point-clouds is mainly caused by the error in translations estimation rather than rotation estimation. The exact implementation is included in our code release.

\smallskip \noindent \textbf{Mapper---Custom Weight Update Rule for a 2D Occupancy Map:} First, we initialize each cell of the 2D occupancy with a weight of 0. For each point in the 3D cloud classified as a free space, we discount the corresponding cell weight on the 2D Occupancy Map by a factor of 0.9. For each point in the 3D cloud classified an obstacle, we add weight $w$ to the corresponding cell on the 2D Occupancy Map, where $w$ is calculated as $w=\frac{0.5}{d}$, and $d$ is the distance between the agent's viewpoint and a point in the 3D cloud. This weighting essentially assigns a lower confidence score for an obstacle on the 2D Occupancy Map which is farther from the current viewpoint. This is due to to the fact that mapping from a pixel in a depth image to a cell in the 2D occupancy map gets more imprecise when $d$ is larger (especially given a depth image in MINOS is coarse.) For an obstacle classified by the collision mapper, we assign weight 1 to the corresponding cell and weights between 0.1 and 0.5 to its neighbors, as detecting the location of an obstacle from a collision is not always precise. Finally, we apply the interpolation to the Occupancy map and cap weight in each cell between 0 and 1.

\smallskip \noindent \textbf{Planner---Custom Edge Weight Calculation Rule:} As described in the paper, we convert the 2D occupancy map into a directed graph; a cell in the map corresponds to a node in the graph. We calculate the weight of an edge from node A to B by taking the sum of (1) the weight of a cell B multiplied by 2000 and (2) a fixed edge cost--1.4 if the edge is diagonal, otherwise 1. 

\subsection* {A2. Classical Navigation Pipeline in S3DIS}~\label{sec:A2}
 In the previous sections, we described the module level differences between the classical navigation pipeline in S3DIS and the pipeline in MINOS. Fig.~\ref{fig:minos_classical_navigation_pipeline} and~\ref{fig:s3dis_classical_navigation_pipeline} illustrate the difference between two versions. In this section, we provide additional details 
 regarding the difference between their mappers.

\smallskip \noindent \textbf{The Differences of Mapper:} The critical difference of mappers in S3DIS and MINOS implementation is that the S3DIS implementation keeps two 2D occupancy maps while MINOS implementation keeps only one 2D occupancy map. The purpose of having multiple 2D occupancy maps is to improve the precision of traverse prediction based on the traverse direction. Intuitively, it is possible to have a cell on 2D occupancy map which may be traversed vertically but not horizontally and vice versa. In S3DIS, given that only two traverse directions (horizontal or vertical) are possible for the agents, we found it beneficial to track the traversal for each of the two possible directions in separate 2D occupancy maps. While keeping multiple 2D occupancy maps may also be beneficial in MINOS, we avoid such an implementation as the number of 2D occupancy maps grows infinite for the continuous traverse directions possible in MINOS.    

The formulation of cells in the 2D occupancy maps are also slightly different across MINOS and S3DIS implementations. In the previous section, we explained that each cell of the 2D occupancy map in MINOS holds a continuously [0,1] valued score which reflects the system's confidence that it is an obstacle. However, in S3DIS, the confidence score stored in each cell is binary. This difference is due to the fact that mapping from a pixel in a depth image to a cell in the 2D occupancy map can be done more precisely in S3DIS because depth images in S3DIS are significantly finer than those in MINOS.

\subsection* {A3. Implementation Details of Idealized 2D Monte Carlo (2D-MC) Agent }~\label{sec:A3}
In this section, we describe the implementation details of the idealized 2D-Monte Carlo (2D-MC) agent. As explained in section 8.1 of the paper, 2D-MC follows four stages in its exploration: (a) perception of visible free space and obstacles, (b) calculation of frontiers, (c) sampling of a subgoal and (d) action planning. Fig.~\ref{fig:2d-mc_navigation_pipeline} illustrates our 2D-MC pipeline.

First, 2D-MC perceives visible free space and obstacles. The goal of this stage is to label each cell in the ground truth 2D occupancy map into one of three classes---free space, obstacle and unobserved. In our implementation, we use a ground truth 2D occupancy map and a wall map (showing the location of walls) to calculate visible regions. We assume that 2D-MC observes all free space and obstacles in all directions in the ground truth 2D map, but cannot see through walls. In MINOS, we obtain both the ground-truth occupancy map and wall map from the room configuration information in the SUNCG dataset. In S3DIS, we obtain a ground-truth occupancy map from floor regions in each scene of S3DIS, and we annotate a wall map. In Fig.~\ref{fig:2d-mc_navigation_pipeline}, red areas on the maps illustrate visible free space.  

Second, we define a frontier to be the boundary between observed free space and unobserved space. A frontier is a candidate of a subgoal which might be on the optimal path to the goal. In Fig.~\ref{fig:2d-mc_navigation_pipeline}, light green regions illustrate the frontiers. 

Third, we choose a subgoal from the frontiers. However, if 2D-MC observes the goal directly, it plans actions toward the goal and terminates an episode. Otherwise, 2D-MC randomly chooses a subgoal from the frontiers. We reduce the probability of a cluttered frontier getting selected as a subgoal by dividing the sample probability by the number of other frontiers in its 5 X 5 neighborhood. The intuition behind this is we want 2D-MC to have an equal chance to explore both wide and narrow openings (e.g., door). Note that we keep the subgoal constant until it is reached by 2D-MC (i.e., we will skip this frontier selection stage and the previous frontier generation stage unless we reach the subgoal.) 

Finally, we perform a path planning and calculate the next action based on the visible free space, visible obstacles, and subgoal. 2D-MC uses A-star search to calculate a path which only traverses through the visible tree space. In S3DIS, we set the action space of 2D-MC to be the same as the original task. In MINOS, we keep the rotation actions as they were in the original task. However, we made the step size of translation action to half of the value it was in the original task because the cell size of the ground truth map is the half of the step size of an agent in MINOS. Therefore, a translation action of 2D-MC in MINOS costs 0.5 steps. We ignore physics in the environment. The final action is chosen so that it approximates the closest orientation to the next cell on the path. If the closest orientation is already achieved, the agent takes an action to translate to the next cell.

{\small
\bibliographystyle{ieee}
\bibliography{ms}
}

\end{document}